# Experiments on Paraphrase Identification Using Quora Question Pairs Dataset

**Andreas Chandra**[*]  **Ruben Stefanus**[*]

Jakarta Artificial Intelligence Research, Jakarta, Indonesia

`{andestjen, rstefanus16}@gmail.com`

**Abstract**

We modeled the Quora question pairs dataset to identify a similar question. The dataset that we use is provided by Quora. The task is a binary classification. We tried several methods and algorithms and different approach from previous works. For feature extraction, we used Bag of Words including Count Vectorizer, and Term Frequency-Inverse Document Frequency with unigram for XGBoost and CatBoost. Furthermore, we also experimented with WordPiece tokenizer which improves the model performance significantly. We achieved up to 97 percent accuracy. Code and Dataset[1].

## 1 Introduction

Paraphrase Identification is a task where a model should identify whether a pair of sentences or documents is a paraphrase. In a business context, paraphrase identification is significantly helpful to improve the user experience of the platform. For instance, in a question-answer forum such as Quora, the user may have the same question that has been asked and answered. Therefore, before the user create a post, there is a pop-up suggestion for the similar question. Furthermore, paraphrase identification is also used in many natural language processing applications such as machine translation, information retrieval, answer selection question and answering system, and more.

There is extensive research on paraphrase identification. There are a lot of methods and approaches to solve this task. Many previous researches used deep learning for classifier algorithms such as Multi-layer Perceptron (Chen et al., 2019; Shi et al., 2019), Convolutional Neural Network to create vector or classifier (Yin and Schütze, 2015; Bonadiman et al., 2019) and combined with attention layer (Yin et al., 2016), or build a classifier with Long Short-Term Memory (Hunt et al., 2019; Adouane et al., 2019; Wieting and Gimpel, 2018; Wang et al., 2017) Bidirectional-LSTM (Wieting et al., 2019) and recent work using Bi-LSTM with Attention (Duong et al., 2019) and more sophisticated model using Transformer which utilized encoder-decoder (Gan and Ng, 2019). In recent years, most of the work focus on paraphrase identification on multilingual, however, there is also considered to solve on cross-lingual paraphrase identification (Alzahrani and Aljuaid, 2020; Yang et al., 2019)

We also found numerous recent works on sematic text similarity. The goal of the task is the same as paraphrase identification, to decide how identical of two pairs of sentences. However, the approach is different from paraphrase identification. Text similarity relies on similar metrics such as Cosine similarity and other various similarity metrics and the result is range from 0 to 1. Therefore, to decide similar or not you have to set a threshold for the score (Fernando and Stevenson, 2008; Neves et al., 2019; Kim et al., 2019; He et al., 2015; Hussain and Suryani, 2015).

Recently, the trend of paraphrase identification has been changing. We see recent papers more focus on paraphrase generation (Qian et al., 2020; Gu et al., 2019; Colin and Gardent, 2020; Li et al., 2018; Witteveen and Andrews, 2019; Yang et al., 2019; Egonmwan and Chali, 2019; Yasui et al., 2019) instead of paraphrase identification, due to the reliability of a model for detecting paraphrase sentence and more challenging task.

However, although previous work has been giving promising results, we would like to achieve a higher score than existing approaches and algorithms, and we found that there are a few studies that examine on Bag of Words and concatenate pairs of sentence and feed in classifier algorithms.

---
[*]equal contribution
[1] https://github.com/jakartaresearch/quora-question-pairs

In this study, we examine Quora Question Pairs dataset using Bag of Words (BoW) and Word Piece to chunk the text and also using tree boosting algorithm that already widely used such as Catboost, XGBoost. Moreover, we also conducted a study on deep learning algorithms such as LSTM and BERT to see how effective to classify paraphrase identification.

The present paper is organized as follows: Section 2 presents the Related Work that inspires this study, Section 3 describes the algorithms for the classifier, Section 4 shows the experiment that had been conducted, and finally, Section 5 is the Conclusion of this study.

## 2 Related Work

In the past decades, there was a growing trend in the field of paraphrase identification. We saw that many datasets and challenges had been made for this task. Even, ACL created a leaderboard to find the state of the art for current result by using Microsoft Research Paraphrase Corpus as a dataset[2]. However, in this section, we focus on previous work that examined Quora Question pairs and we can compare the previous result with the present paper result.

(Peinelt et al., 2020) conducted studies on many datasets such as SemEval CQA, MSRP, Quora, STS Benchmark dataset to find the obvious and non-obvious sentence pair. The study finds that pairs with high word divergence tend to have a negative label on the observed datasets, while low word divergence is associated with positive label and Quora dataset that we used in this study has more obvious sentence pair instead of non-obvious. The study also proposed a criterion for evaluation which more emphasize on true positive rate and true negative rate on obvious and non-obvious sentence pair.

(Wang et al., 2017) proposed a new model namely bilateral multi-perspective matching (BiMPM) which utilizes BiLSTM as an encoder and another BiLSTM layer to aggregate the matching result into a fix-length matching vector and continued to last layer of the model which is a fully connected layer. The paper use GloVe as a pretrained word vector from 840B Common Crawl corpus and apply it to Quora Question Pairs and select random sample from it. The paper also benchmarks with other approaches, and the proposed model, BiMPM gives a stunning result that achieves 88.8 percent accuracy.

(Bonadiman et al., 2019) train dataset using a set of layers that involve embeddings, convolutional, and global max pooling. The aim of the research is to give the k-best similar questions that may have been answered given a query input to the community question answering platform. Moreover, the paper also proposed a new criterion, smoothed deep metric, to optimize the model and model as a classification problem instead classify a given query compare to all other questions in the dataset.

(Shi et al., 2019) proposed a new approach that is rare compared to the other recent studies. The proposed model utilized contextual embeddings, ELMo, instead of a random embedding in the beginning. This is reasonable because contextual embedding gives a meaningful representation of words and word's context. However, the paper argues that contextualized embedding of a token significantly changes and may shift the representation drastically when the context is paraphrased. The proposed model tested with three different well-known datasets, Microsoft Research Paraphrase Corpus, Sampled Quora, and PAN. It shows a better result compared to the original ELMo.

## 3 Methodology

The present paper, we tried various feature extractions and algorithms which are widely used for classification task.

**Feature extraction**, after the data is clean, we extract the sentence to get the matrix. we experimented using Bag of Words (BoW) for XGBoost (Chen and Guestrin, 2016) and Catboost (Prokhorenkova et al., 2018). Word Piece using tokenizers library provided by HuggingFace[3] for BERT (Devlin et al., 2019) and LSTM (Hochreiter and Schmidhuber, 1997) model. We used NLTK[4] word tokenizer for building term frequency dictionary. We experimented with both unigram Count Vectorizer (CV) and Term Frequency-Inverse Document Frequency (TF-IDF). We also tried with Word2Vec and Doc2Vec by Gensim, but the result is weird, it seemed the gradient was not changing and stuck at 66 percent accuracy, and we do not proceed with the experiment using pretrained embedding.

---

[2] https://aclweb.org/aclwiki/Paraphrase_Identification_(State_of_the_art)
[3] https://github.com/huggingface/tokenizers
[4] https://www.nltk.org/

**Model,** we benchmark popular algorithms which mentioned above. We consider the boosting algorithm and categorical as the main advantage of the algorithm as in the token as the word that occurred in the corpus. Both CatBoost and XGBoost used default parameters for the fitting. We utilized LSTM and BERT as a classifier model because it is widely used for sequence classification. The more detailed configuration and settings of the training phase are explained in section 4.

## 4 Experiments

In this section, we describe the dataset, data preprocessing, and configuration of the model.

The dataset consists of 404,290 observations and each observation consists of question 1 and question 2 to be paired with binary label, as 1 to indicate duplicate and 0 to indicate otherwise. Quora Question Pairs has more negative label, dominate about 64 percent of the corpus. We still consider the data as a balanced dataset because it is not very significant to take it to the imbalance problem. We did not oversample or under-sampling for it. We also created partitions of the corpus and slash into train and test set using Stratified Cross Validation using scikit-learn[5] with 5 folds. We did 5 cross-validations to the Catboost and XGBoost. However, we only test the first fold for LSTM and BERT. We also tested BERT using test set provided by (Wang et al., 2017) to find out how effective BERT to this task.

There are two steps before the data fed into the model. One is the preprocessing step which consists of removing stop words and unnecessary characters, then in the feature extraction step, the sentence was tokenized and transform to count-vectorizer and term frequency-inverse document frequency for XGBoost and CatBoost model. Because this task has a pair of questions, then the feature extraction of each question1 and question2 have its own matrix.

However, BERT and LSTM Model, we used Word Piece model as a tokenizer and no preprocessing needed for these experiments. We found out that word piece tokenizer actually eliminates unnecessary character and count the most occurrence sub word in the corpus.

We set the parameters of LSTM model as simple as possible. We set hidden size to 512 for Embedding, 512 for hidden size and 2 layers LSTM and set dropout at 0.5 to reduce overfitting, and the last we set number of class unit for Linear layer. We used Adam optimizer with learning rate = 0.001, and Cross Entropy as a loss function.

The configurations of BERT are as follows: We use BERT-Base-Cased model and add a single linear layer on top for classification. Cased means that the true case and accent markers are preserved, and case information is important in this task. Then, for the BERT-Base architecture, it consists of transformer blocks denoted as L, hidden size as H and self-attention heads as A. Then, we use a common setting for BERT as in L=12, H=768, A=12, as a result, the total trainable parameters are 110 Million.

The maximum sequence length we used is 64 tokens, then for the size of batch data, we used 32. Optimizer which we used is Adam algorithm with weight decay fix and the following parameters, learning rate = 2e-5 and epsilon = 1e-8. We also used learning rate scheduler which will make learning rate value decreases linearly. And we did scale for the gradients down to maximum norm in 1, this technique we did to help prevent the exploding gradients problems. Finally, BERT model is trained for 3 epochs.

| Methods | Accuracy |
| --- | --- |
| CV-XGBoost | 68.09 |
| CV-CatBoost | 74.66 |
| TF-IDF-XGBoost | 69.14 |
| TF-IDF CatBoost | **75.39** |

Table 1: The mean accuracy of Xgboost and Catboost on difference feature extraction with 5 Folds

Table 1 shows that boosting algorithms only give the best result up to around 75 percent accuracy for average accuracy from 5 folds. Firstly, we began this experiment to see the baseline and not suddenly jump to deep neural network. Secondly, we would like test if the concatenation of bag of word vector is really work. It is needed to justify that we need experiment on deep learning. However, accuracy might not the best metrics for the task. because, we also calculated and see on precision and recall score, and the result it not that well. The f-score is only 59.25 percent for TF-IDF Catboost.

---

[5] https://scikit-learn.org/

| Methods | Accuracy |
| --- | --- |
| LSTM | 80.32 |
| BERT | 97.07 |

Table 2: accuracy of LSTM and BERT

Table 2 displays the result of LSTM and BERT. We already expect that BERT will give a stunning result given that it consists of sophisticated layers and very deep network for the task.

| Methods | Accuracy |
| --- | --- |
| BERT | 97.08 |
| Zhiguo, et al | 88.8 |

Table 3: a comparison previous work and present paper using BERT

We finally compared BERT original model and model proposed by (Wang et al., 2017) in Table 3 and using their dataset to infer the same test data, then we can compare the result. It shows that BERT is still the best.

## 5 Conclusion

In this work, we experimented Bag of Words with boosting algorithms, Catboost, and XGBoost. We also tested Quora Question Pairs using simple LSTM and BERT. The results show that BERT gave the best result among other models. For further research, we suggest that more work on paraphrase generation which may be more significant to the natural language processing task such as machine translation and question answering system.

## Acknowledgement

We gratefully acknowledge the support of Jakarta Artificial Intelligence Research to provide the facilities to conduct the research. We also would like to thank Andhika Setia Pratama for proof reading and helpful comments.

## References

Wafia Adouane, Jean-Philippe Bernardy, and Simon Dobnik. 2019. Neural Models for Detecting Binary Semantic Textual Similarity for Algerian and MSA. In *Proceedings of the Fourth Arabic Natural Language Processing Workshop*, number 2, pages 78–87, Stroudsburg, PA, USA. Association for Computational Linguistics.

Salha Alzahrani and Hanan Aljuaid. 2020. Identifying cross-lingual plagiarism using rich semantic features and deep neural networks: A study on Arabic-English plagiarism cases. *Journal of King Saud University - Computer and Information Sciences*(xxxx), April.

Daniele Bonadiman, Anjishnu Kumar, and Arpit Mittal. 2019. Large Scale Question Paraphrase Retrieval with Smoothed Deep Metric Learning. In *Proceedings of the 5th Workshop on Noisy User-generated Text (W-NUT 2019)*, pages 68–75, Stroudsburg, PA, USA. Association for Computational Linguistics.

Tianqi Chen and Carlos Guestrin. 2016. XGBoost. In *Proceedings of the 22nd ACM SIGKDD International Conference on Knowledge Discovery and Data Mining*, volumes 13-17-Augu, pages 785–794, New York, NY, USA, August. ACM.

Weize Chen, Hao Zhu, Xu Han, Zhiyuan Liu, and Maosong Sun. 2019. Quantifying Similarity between Relations with Fact Distribution. In *Proceedings of the 57th Annual Meeting of the Association for Computational Linguistics*, pages 2882–2894, Stroudsburg, PA, USA. Association for Computational Linguistics.

Emilie Colin and Claire Gardent. 2020. Generating syntactic paraphrases. In *Proceedings of the 2018 Conference on Empirical Methods in Natural Language Processing, EMNLP 2018*, pages 937–943, Stroudsburg, PA, USA. Association for Computational Linguistics.

Jacob Devlin, Ming-Wei Chang, Kenton Lee, and Kristina Toutanova. 2019. BERT: Pre-training of Deep Bidirectional Transformers for Language Understanding. In *Proceedings of the 2019 Conference of the North*, pages 4171–4186, Stroudsburg, PA, USA, October. Association for Computational Linguistics.

Phuc H. Duong, Hien T. Nguyen, Hieu N. Duong, Khoa Ngo, and Dat Ngo. 2019. A Hybrid Approach to Paraphrase Detection. In *NICS 2018 - Proceedings of 2018 5th NAFOSTED Conference on Information and Computer Science*, number i, pages 366–371. IEEE, November.

Elozino Egonmwan and Yllias Chali. 2019. Transformer and seq2seq model for Paraphrase Generation. In *Proceedings of the 3rd Workshop on Neural Generation and Translation*, number Wngt, pages 249–255, Stroudsburg, PA, USA. Association for Computational Linguistics.

Samuel Fernando and Mark Stevenson. 2008. A Semantic Similarity Approach to Paraphrase Detection. *Proceedings of the 11th Annual Research Colloquium of the UK Special Interest Group for Computational Linguistics (CLUK 2008)*:45–52.

Wee Chung Gan and Hwee Tou Ng. 2019. Improving the Robustness of Question Answering Systems to Question Paraphrasing. In *Proceedings of the 57th Annual Meeting of the Association for Computational Linguistics*, pages 6065–6075, Stroudsburg, PA, USA. Association for Computational Linguistics.

Yunfan Gu, yang yuqiao, and Zhongyu Wei. 2019. Extract, Transform and Filling: A Pipeline Model for Question Paraphrasing based on Template. In *Proceedings of the 5th Workshop on Noisy User-generated Text (W-NUT 2019)*, pages 109–114, Stroudsburg, PA, USA. Association for Computational Linguistics.

Hua He, Kevin Gimpel, and Jimmy Lin. 2015. Multi-Perspective Sentence Similarity Modeling with Convolutional Neural Networks. In *Proceedings of the 2015 Conference on Empirical Methods in Natural Language Processing*, number September, pages 1576–1586, Stroudsburg, PA, USA. Association for Computational Linguistics.

Sepp Hochreiter and Jürgen Schmidhuber. 1997. Long Short-Term Memory. *Neural Computation*, 9(8):1735–1780, November.

Ethan Hunt, Binay Dahal, Justin Zhan, Laxmi Gewali, Paul Oh, Ritvik Janamsetty, Chanana Kinares, Chanel Koh, Alexis Sanchez, Felix Zhan, Murat Ozdemir, Shabnam Waseem, and Osman Yolcu. 2019. Machine Learning Models for Paraphrase Identification and its Applications on Plagiarism Detection. In *2019 IEEE International Conference on Big Knowledge (ICBK)*, pages 97–104. IEEE, November.

Syed Fawad Hussain and Asif Suryani. 2015. On retrieving intelligently plagiarized documents using semantic similarity. *Engineering Applications of Artificial Intelligence*, 45:246–258, October.

Yunsu Kim, Hendrik Rosendahl, Nick Rossenbach, Jan Rosendahl, Shahram Khadivi, and Hermann Ney. 2019. Learning Bilingual Sentence Embeddings via Autoencoding and Computing Similarities with a Multilayer Perceptron. In *Proceedings of the 4th Workshop on Representation Learning for NLP (RepL4NLP-2019)*, pages 61–71, Stroudsburg, PA, USA. Association for Computational Linguistics.

Zichao Li, Xin Jiang, Lifeng Shang, and Hang Li. 2018. Paraphrase Generation with Deep Reinforcement Learning. In *Proceedings of the 2018 Conference on Empirical Methods in Natural Language Processing*, pages 3865–3878, Stroudsburg, PA, USA. Association for Computational Linguistics.

Mariana Neves, Daniel Butzke, and Barbara Grune. 2019. Evaluation of Scientific Elements for Text Similarity in Biomedical Publications. In *Proceedings of the 6th Workshop on Argument Mining*, number c, pages 124–135, Stroudsburg, PA, USA. Association for Computational Linguistics.

Nicole Peinelt, Maria Liakata, and Dong Nguyen. 2020. Aiming beyond the obvious: Identifying non-obvious cases in semantic similarity datasets. In *ACL 2019 - 57th Annual Meeting of the Association for Computational Linguistics, Proceedings of the Conference*, pages 2792–2798, Stroudsburg, PA, USA. Association for Computational Linguistics.

Liudmila Prokhorenkova, Gleb Gusev, Aleksandr Vorobev, Anna Veronika Dorogush, and Andrey Gulin. 2018. Catboost: Unbiased boosting with categorical features. In *Advances in Neural Information Processing Systems*, volumes 2018-Decem, pages 6638–6648.

Lihua Qian, Lin Qiu, Weinan Zhang, Xin Jiang, and Yong Yu. 2020. Exploring diverse expressions for paraphrase generation. In *EMNLP-IJCNLP 2019 - 2019 Conference on Empirical Methods in Natural Language Processing and 9th International Joint Conference on Natural Language Processing, Proceedings of the Conference*, pages 3173–3182, Stroudsburg, PA, USA. Association for Computational Linguistics.

Weijia Shi, Muhao Chen, Pei Zhou, and Kai-Wei Chang. 2019. Retrofitting Contextualized Word Embeddings with Paraphrases. In *Proceedings of the 2019 Conference on Empirical Methods in Natural Language Processing and the 9th International Joint Conference on Natural Language Processing (EMNLP-IJCNLP)*, pages 1198–1203, Stroudsburg, PA, USA. Association for Computational Linguistics.

Zhiguo Wang, Wael Hamza, and Radu Florian. 2017. Bilateral multi-perspective matching for natural language sentences. In *IJCAI International Joint*

*Conference on Artificial Intelligence*, pages 4144–4150, California, August. International Joint Conferences on Artificial Intelligence Organization.

John Wieting and Kevin Gimpel. 2018. ParaNMT-50M: Pushing the Limits of Paraphrastic Sentence Embeddings with Millions of Machine Translations. In *Proceedings of the 56th Annual Meeting of the Association for Computational Linguistics (Volume 1: Long Papers)*, volume 1, pages 451–462, Stroudsburg, PA, USA. Association for Computational Linguistics.

John Wieting, Kevin Gimpel, Graham Neubig, and Taylor Berg-Kirkpatrick. 2019. Simple and Effective Paraphrastic Similarity from Parallel Translations. In *Proceedings of the 57th Annual Meeting of the Association for Computational Linguistics*, pages 4602–4608, Stroudsburg, PA, USA. Association for Computational Linguistics.

Sam Witteveen and Martin Andrews. 2019. Paraphrasing with Large Language Models. In *Proceedings of the 3rd Workshop on Neural Generation and Translation*, number Wngt, pages 215–220, Stroudsburg, PA, USA. Association for Computational Linguistics.

Yinfei Yang, Yuan Zhang, Chris Tar, and Jason Baldridge. 2019. PAWS-X: A Cross-lingual Adversarial Dataset for Paraphrase Identification. In *Proceedings of the 2019 Conference on Empirical Methods in Natural Language Processing and the 9th International Joint Conference on Natural Language Processing (EMNLP-IJCNLP)*, pages 3685–3690, Stroudsburg, PA, USA. Association for Computational Linguistics.

Go Yasui, Yoshimasa Tsuruoka, and Masaaki Nagata. 2019. Using Semantic Similarity as Reward for Reinforcement Learning in Sentence Generation. In *Proceedings of the 57th Annual Meeting of the Association for Computational Linguistics: Student Research Workshop*, pages 400–406, Stroudsburg, PA, USA. Association for Computational Linguistics.

Wenpeng Yin and Hinrich Schütze. 2015. Convolutional neural network for paraphrase identification. In *NAACL HLT 2015 - 2015 Conference of the North American Chapter of the Association for Computational Linguistics: Human Language Technologies, Proceedings of the Conference*, pages 901–911, Stroudsburg, PA, USA. Association for Computational Linguistics.

Wenpeng Yin, Hinrich Schütze, Bing Xiang, and Bowen Zhou. 2016. ABCNN: Attention-Based Convolutional Neural Network for Modeling Sentence Pairs. *Transactions of the Association for Computational Linguistics*, 4:259–272, December.